\pdfoutput=1
\PassOptionsToPackage{table}{xcolor}
\documentclass[11pt]{article}

\usepackage{authblk}

\usepackage[]{EMNLP2023}

\usepackage{draftwatermark}
\SetWatermarkText{Pre-print}
\SetWatermarkLightness{0.9}

\usepackage[normalem]{ulem}
\useunder{\uline}{\ulined}{}

\usepackage{times}
\usepackage{latexsym}

\usepackage[T1]{fontenc}
\usepackage[utf8]{inputenc}

\usepackage{microtype}

\usepackage{inconsolata}

\usepackage{amsmath}
\usepackage{amssymb}
\usepackage{color}
\usepackage{soul}

\usepackage{hyperref}

\usepackage{caption}
\usepackage{subcaption}

\usepackage{booktabs}
\usepackage{multirow}

\usepackage{ bbold }

\usepackage{textcomp}

\usepackage{graphicx}
\graphicspath{%
    {figures/}
}

\usepackage[nameinlink]{cleveref}

\hypersetup{
  colorlinks = true, %Colours links instead of ugly boxes
  urlcolor = blue, %Colour for external hyperlinks
  linkcolor = blue, %Colour of internal links
  citecolor = red %Colour of citations
}

\newcommand{\blind}[2]{\ifdefined\blinded#2\else#1\fi}

\newcommand{\ie}{\textit{i.e.}}             % literally
\newcommand{\eg}{\textit{e.g.}}             % For example
\title{%
    Measuring Misogyny in Natural Language Generation:\\
    Preliminary Results from a Case Study on two Reddit Communities%
}

\author{%
\bf
Aaron~J.~Snoswell$^{1,2,4}$, 
Lucinda~Nelson$^{1,3,4}$,  
Hao~Xue$^{1,5}$,   
Flora~D.~Salim$^{1,5}$\\
\bf 
Nicolas~Suzor$^{1,3,4}$, 
Jean~Burgess$^{1,2,4}$\vspace{-1em} \\ \\
\small $^1$Australian Research Council Centre of Excellence for Automated Decision-Making and Society \vspace{-0.3em} \\
\small \{$^2$School of Communications, $^3$School of Law, $^4$Digital Media Research Centre\}, \vspace{-0.3em} \\
\small Queensland University of Technology, Brisbane, QLD, Australia \vspace{-0.3em} \\
\small \texttt{\{%
    \href{mailto:a.snoswell@qut.edu.au}{a.snoswell},%
    \href{mailto:l5.nelson@qut.edu.au}{l5.nelson},%
    \href{mailto:n.suzor@qut.edu.au}{n.suzor},%
    \href{mailto:jean.burgess@qut.edu.au}{jean.burgess}%
\}@qut.edu.au} \vspace{-0.3em} \\
\small $^5$School of Computer Science and Engineering, University of New South Wales, Sydney, NSW, Australia \vspace{-0.3em} \\
\small \texttt{\{%
    \href{mailto:hao.xue1@unsw.edu.au}{hao.xue1},%
    \href{mailto:flora.salim@unsw.edu.au}{flora.salim}%
\}@unsw.edu.au}
}

\newcommand{\ric}{\textit{r/Incels}}
\newcommand{\rfa}{\textit{r/ForeverAlone}}

\makeatletter
\def\blfootnote{\xdef\@thefnmark{}\@footnotetext}
\makeatother

\begin{document}
\maketitle
\begin{abstract}

Generic `toxicity' classifiers continue to be used for evaluating the potential for harm in natural language generation, despite mounting evidence of their shortcomings.
We consider the challenge of measuring misogyny in natural language generation, and argue that generic `toxicity' classifiers are inadequate for this task.
We use data from two well-characterised `Incel' communities on Reddit that differ primarily in their degrees of misogyny to construct a pair of training corpora which we use to fine-tune two language models.
We show that an open source `toxicity' classifier is unable to distinguish meaningfully between generations from these models.
We contrast this with a misogyny-specific lexicon recently proposed by feminist subject-matter experts, demonstrating that, despite the limitations of simple lexicon-based approaches, this shows promise as a benchmark to evaluate language models for misogyny, and that it is sensitive enough to reveal the known differences in these Reddit communities.
Our preliminary findings highlight the limitations of a generic approach to evaluating harms, and further emphasise the need for careful benchmark design and selection in natural language evaluation.

\end{abstract}

\blfootnote{
    This extended abstract was presented at the Generation, Evaluation and Metrics workshop at Empirical Methods in Natural Language Processing in 2023 (GEM@EMNLP 2023) in Singapore.
}

\section{Introduction and Related Work}

The measurement of potentially harmful generation tendencies in language models is an important topic which has recently been studied in great detail.
Yet despite many documented shortcomings (see \eg{} \Cref{app:shortcomings-of-perspective-api}), generic `toxicity' classifiers such as Jigsaw's Perspective API continue to be a \textit{de-facto} standard for operationalising and benchmarking harmful generation tendencies -- for instance, forming a key part of the RealToxicityPrompts \cite{GehmanRealToxicityPrompts2020}, CivilComments \cite{DucheneBenchmark2023} and HELM \cite{BommasaniHolistic2023} evaluation frameworks.

Recent work has highlighted the hazards that black-box APIs pose by hampering reproducible natural language processing research \cite{PozzobonChallenges2023}.
Here, we shed light on a different dimension of this problem: the inadequacy of generic `toxicity' classifiers to detect characteristics of harmful expression associated with specific ideologies and counterspeech, such as misogyny and counter-misogyny.

% \usepackage{booktabs}
% \usepackage{multirow}
% \usepackage{graphicx}
\begin{table}[t]
\caption{%
    Misogyny evaluation results on RealToxicityPrompts dataset, with $k=25$ completions per prompt.
    According to an open-source `toxicity' classifier (Detoxify-multilingual), a language model fine-tuned on posts from the \ric{} community appears worse, or equally as bad, as a similar model fine-tuned on posts from the \rfa{} community.
    However, a simple lexicon approach developed by subject matter experts (the Farrell lexicon) shows the opposite result, aligning with extensive prior qualitative work by subject-matter experts characterizing these communities.
    EML - Expected Maximum Likelihood, BFC - Binary Classifier Frequency.
    Values are mean percentage and 95\% confidence intervals, lower is better.
}
\label{tab:main-results}
\vspace{-6pt}
\resizebox{\columnwidth}{!}{%
\begin{tabular}{@{}rccc@{}}
\toprule
\multirow{2}{*}{Model} &
  \multicolumn{2}{c}{\begin{tabular}[c]{@{}c@{}}Detoxify-multilingual\\ `toxicity'\end{tabular}} &
  \begin{tabular}[c]{@{}c@{}}Farrell lexicon\\ Any misogyny\end{tabular} \\ \cmidrule(l){2-4} 
             & EML                                & BCF                               & BCF                        \\ \midrule
Pre-trained  & $49.35 \pm 0.24$                   & \textbf{\boldmath$6.65 \pm 0.08$} & $5.72 \pm 0.05$                   \\[5pt]
Incels       & {\cellcolor[HTML]{D9E2F3}} \textbf{\boldmath$35.05 \pm 0.24$} & {\cellcolor[HTML]{FBE4D5}} $12.03 \pm 0.13$                  & {\cellcolor[HTML]{FBE4D5}} $9.40 \pm 0.11$                   \\
ForeverAlone & {\cellcolor[HTML]{FBE4D5}} $76.71 \pm 0.18$                   & {\cellcolor[HTML]{FBE4D5}} $9.08 \pm 0.06$                   & {\cellcolor[HTML]{D9E2F3}} \textbf{\boldmath$5.50 \pm 0.04$} \\ \bottomrule
\end{tabular}%
}
\vspace{-10pt}
\end{table}

As a case study, we use misogyny on the social media site Reddit.
\Citep{ManneGirl2017} defines misogyny as the enforcement of patriarchy.
This can be expressed overtly or covertly, and can be couched in civil language (as in `mansplaining'), or violence.
Extensive feminist scholarship has examined misogyny on Reddit, particularly a cluster of `manosphere' communities on the site \cite{GingAlphas2019,TrottOperationalising2022,HelmExamining2022}.
We build on prior deep qualitative work that has specifically focused on the communities \ric{} and \rfa{} through a feminist lens \cite{GillettIncels2022} -- and which demonstrated that these communities are similar in many respects but differ markedly in the degree of misogynistic ideology they express (see \Cref{app:the-incels-and-foreveralone-subreddit-communities}).

Armed with this knowledge as a ground truth, we use submissions posted to two communities from 2011 to 2016 to construct a pair of probe corpora for fine-tuning to explore measurement of misogynistic generations in language models (\Cref{app:data-collection-and-fine-tuning} details our data collection and fine-tuning methodology).
Prompting with RealToxicityPrompts \cite{GehmanRealToxicityPrompts2020}, we then explore the ability of two benchmarking approaches (the open-source Detoxify-multilingual classifier \cite{HanuDetoxify2020}, and a misogyny-specific lexicon developed by feminist subject-matter experts \cite{FarrellExploring2019}) to differentiate these models.

In summary, the contributions of this paper are to: (i) caution against the use of generic `toxicity' benchmarks as a catch-all for diverse societal harms, and to (ii) demonstrate the potential of a simple lexicon-based approach for LLM generation evaluation, when subject-matter expertise is appropriately engaged in the evaluation process.

\section{Results}

We use the Pythia suite of large language models for our experiments \citep{BidermanPythia2023}, fine-tune on Reddit data captured by the PushShift API project \citep{BaumgartnerPushshift2020}, and prompt the models with the RealToxicityPrompts dataset \citep{GehmanRealToxicityPrompts2020}.
\Cref{app:evaluation-protocol} describes our full evaluation protocol and defines the metrics we report in our main results (\Cref{tab:main-results}).

We find that scores from the Detoxify-multilingual classifier are mixed, providing no clear indication which of the fine-tuned models is better or worse.
On the EML (a measure of worst-case `toxicity' likelihood), the \ric{} model actually performs better than the pre-trained model (which is extremely surprising given the extensive documentation of the toxic behaviour in this community in prior literature), while the BCF suggests both models are worse than the pre-trained state, with \ric{} worse than \rfa{}.
The difficulty distinguishing these models stems from the similarities in the corpora -- and is borne out in the parameter vectors for each model, which were found to be extremely similar in parameter space with a cosine similarity of $0.99985$.

However on the Farrell lexicon scores, we see that the \ric{} model produces more misogynistic generations that the \rfa{}, and the \rfa{} model actually scores better than the pre-trained model.
This aligns with the prior subject-matter expert qualitative scholarship studying these Reddit communities \cite{GillettIncels2022}, which we take to be a source of ground truth.

\section{Discussion and Conclusion}

These results, while preliminary, echo recent scholarship highlighting the importance of careful benchmark design and selection \cite{RauhCharacteristics2022}.
Using only a generic `toxicity' classifier, looking at the EML metric, the \ric{} fine-tuned model receives low `toxcicity' scores.
But considering the Detoxify-multilingual BCF and EML metrics jointly, there is not a clear difference between the \ric{} and \rfa{} fine-tuned models.
These mixed interpretations are also borne out if considering the classifier outputs from this model other than `toxicity' (see \Cref{tab:detoxify-prompted-eml,tab:detoxify-prompted-bcp} in \Cref{app:additional-evaluation-results}).

When considered using the Farrell lexicon though, the result better aligns with the prior literature studying these communities -- \ric{} is seen to be more misogynistic than \rfa{} when looking for any type of misogyny (\Cref{tab:farrell-prompted} in \Cref{app:additional-evaluation-results} breaks down further nuances between these models for the sub-types of misogyny coded in the lexicon).

These preliminary findings suggest that the lexicon-based approaches may have a role to play in the evaluation of LLM generations.
This also demonstrates the limits of generic one-size-fits-all classifiers for harmful model tendencies, and underscores the importance of careful benchmark selection and subject matter expertise in the evaluation of harmful behaviours in natural language generation.

\section*{Limitations}
% EMNLP 2023 requires all submissions to have a section titled ``Limitations'', for discussing the limitations of the paper as a complement to the discussion of strengths in the main text. This section should occur after the conclusion, but before the references. It will not count towards the page limit.  

% The discussion of limitations is mandatory. Papers without a limitation section will be desk-rejected without review.
% ARR-reviewed papers that did not include ``Limitations'' section in their prior submission, should submit a PDF with such a section together with their EMNLP 2023 submission.

% While we are open to different types of limitations, just mentioning that a set of results have been shown for English only probably does not reflect what we expect. 
% Mentioning that the method works mostly for languages with limited morphology, like English, is a much better alternative.
% In addition, limitations such as low scalability to long text, the requirement of large GPU resources, or other things that inspire crucial further investigation are welcome.

These are preliminary results with limited conclusions.
One limitation is that we could not use the Perspective API for `toxicity' classification -- because we were unable to secure a quota increase over the free quota of one query per second.
To run our suite of evaluations over the RealToxicityPrompts dataset would have been prohibitively slow.
We chose not to sub-sample our evaluations and reduce the reliability of our reported results (as is common practice), instead substituting with a similar but open-source classifier.
Another limitation is that we have not yet performed a rigorous qualitative analysis of model generations to go alongside our quantitative analysis.
This work is under way, and will be reported in follow-on work.

\section*{Ethics Statement}
% Scientific work published at EMNLP 2023 must comply with the \href{https://www.aclweb.org/portal/content/acl-code-ethics}{ACL Ethics Policy}. We encourage all authors to include an explicit ethics statement on the broader impact of the work, or other ethical considerations after the conclusion but before the references. The ethics statement will not count toward the page limit (8 pages for long, 4 pages for short papers).

\blind{%
    This study was approved by the QUT Human Research Ethics Application board, approval number 2023-5627-13008 and variation HE-VR-2023-5627-12909.
}{
    This study was approved by the [removed for blind peer review] Human Research Ethics Application board, approval number [removed for blind peer review] and variation [removed for blind peer review].
}

\section*{Acknowledgements}

\blind{%
    This research was funded by the Australian Research Council Centre of Excellence for Automated Decision-Making and Society (CE200100005).
    Computational resources and services used in this work were provided by the eResearch Office, Queensland University of Technology, Brisbane, Australia.
    We also thank our colleagues and the anonymous peer reviewers who provided valuable feedback on this manuscript during the drafting process.
}{%
    Acknowledgements removed for blind peer-review.
}

\bibliography{anthology,bib-EMNLP-2023}
\bibliographystyle{acl_natbib}

\clearpage

\appendix
\onecolumn

\section{Shortcomings of the Perspective API}
\label{app:shortcomings-of-perspective-api}

A full review of the Perspective API is beyond the scope of this article, however we briefly mention some key findings from the literature.
For example, the Perspective classifier is known to correlate strongly with the presence of profanity \cite{JiangCritical2020}, to struggle with subtler expressions of hate \cite{HanFortifying2020}, and to display a host of unintended biases against marginalised communities \cite{RottgerHateCheck2020,MendelsohnDogwhistles2023,ThiagoFighting2021,SapRisk2019,ReichertReading2020} despite dedicated attempts to remove these biases \cite{BorkanNuanced2019}.
Recent work by \citep{TrottOperationalising2022} and \citep{StrathernIdentifying2022} also found that Perspective API specifically struggles with identification of misogynistic ideologies in text.

\section{The \ric{} and \rfa{} subreddit communities}
\label{app:the-incels-and-foreveralone-subreddit-communities}

\ric{}, which stands for `involuntary celibate', was a subreddit dedicated to men sharing their misogynistic feelings of `aggrieved entitlement' \cite{KimmelAngry2017} from 2013 until it was banned by Reddit in 2017.
During this time, \ric{} made news headlines due to its connections with several mass murders, and feminist scholarship has extensively documented the extremely harmful behaviour exhibited by this community.

\rfa{} is a subreddit that is dedicated to people (in particular, men) discussing experiences of loneliness and relationship challenges. Despite covering similar topics as \ric{}, critiques of \rfa{}  have found that the discourse is less misogynistic, not blaming women for men's struggles -- additionally, women are able to participate in this community, and moderators do a better job of controlling the influx or spread of harmful actors and ideologies \cite{GillettIncels2022,MillarPlatforming2022}.

\section{Data collection and fine-tuning}
\label{app:data-collection-and-fine-tuning}

\subsection{Data Collection and Preparation}

We use the PushShift service \cite{BaumgartnerPushshift2020} for accessing archived Reddit data.
We mirror the qualitative analysis of Reddit discourse \cite{GillettIncels2022}, capturing subreddit submissions only during the time window of 2011-2019.
In this period, the PushShift service captured 42,146 and 104,726 submissions from \ric{} and \rfa{} respectively.
We filtered out all posts less than 100 characters, leaving 19,068 and 57,700 items respectively.
After removing duplicates\footnotemark{} with cosine similarity $\ge 0.9$, we had 16,204 and 51,167 posts respectively.
From these, we partition the data into 70/20/10\% train/test/validate sets.

\footnotetext{%
    We use the \href{https://github.com/Bergvca/string_grouper}{StringGrouper python library}.
}

In addition to the above data preparation, we had to perform one  manual filtering step.
There were two submissions in the \ric{} dataset which contained extremely long phrases `REEEEEEEEEEEEE\dots' (continuing for several thousand characters).\footnotemark{}
The presence of multiple entire context windows worth of subsequent `E' characters led to the fine-tuned \ric{} model always choosing `E' as its first token, and gave a very high probability of continuing to generate `EEE' tokens arbitrarily after this.
This disrupted all the model's generations and lead to issues with running any evaluations, so we manually truncated these two long `reee' expressions at 20 characters, then re-trained the models.

\footnotetext{%
    This is an Incel derogotory expression that is meant to imitate a screeching sound.
    It can have ableist or misogynistic connotations (\eg{} in mockery of a screeching, irrational feminist) and is used to shout out new-comers to the forum.
}

\subsection{Fine-tuning}

We use the Pythia suite of open-source decoder-only language models for our experiments \cite{BidermanPythia2023}, which have a 2048-token context window.
All reported results are for the 12B parameter model, which is similar in capability level to the T5 or OPT 6.7B models -- results for smaller models were similar, so we do not include them here.
We fine-tuned the pre-trained model on the \ric{} and \rfa{} datasets using the the HuggingFace transformer library \cite{WolfTransformers2020} with a causal language modelling objective using the AdamW optimizer for 5 epochs and a global batch size of 32.
We use DeepSpeed \cite{RasleyDeepspeed2020} ZeRO-3 offload \cite{RajbhandariZero2020} with 16 gradient accumulation steps and gradient checkpointing, and train using $4\times$ NVidia A100 cards and 512Gb of RAM.

\section{Evaluation protocol}
\label{app:evaluation-protocol}

We use the RealToxicityPrompts dataset \cite{GehmanRealToxicityPrompts2020} to prompt the models, and generate $k=25$ completions per prompt, allowing up to 20 tokens per completion but stopping earlier if the \textit{end-of-sentence} token is generated.
For generation, we use nucleus sampling \cite{HoltzmanCurious2019} with a temperature of 1 and $p=0.9$.

\subsection{Detoxify-multilingual}

We intended to use the popular Jigsaw Perspective API as a `toxicity' classifier baseline, however we were unable to secure a quota-increase above the free tier of one query per second.
In our opinion, this is further evidence of why this tool is inappropriate for academic benchmarking in addition to the excellent points about black-box APIs raised by \citep{PozzobonChallenges2023} and about changes over time in the API raised by \citep{YeNoisyHate2023}.

As a substitute, we used the open-source Detoxify-multilingual classifier to score completions \cite{HanuDetoxify2020}.
This model was developed for a Kaggle competition run by Jigsaw, trained on a sub-set of the Perspective API data, and scored very competitively on the Kaggle competition (see \citep{HanuDetoxify2020}).
These facts, along with our own comparisons of this classifier with Perspective API in \Cref{app:detoxify-perspective-comparison} give us confidence that this classifier is similar to the Perspective API.

Following the recommendations from the RealToxicityPrompts authors \cite{GehmanRealToxicityPrompts2020}, we report the Expected Maximum Likelihood (EML) and Binary Classifier Frequency (BCF) over $k$ completions.

EML is the largest classifier likelihood over the $k$ completions -- a measure of worst-case `toxicity' likelihood;
\begin{align}
    \text{EML} &\triangleq \mathbb{E}_{p \in \mathcal{D}} \left[
        \max_{i \in [1, \dots, k]} \left(
            \sigma(c_i(p))
        \right)
    \right],
\end{align}

\noindent where $c_i$ is the $i^\text{th}$ completion from the model, conditioned on prompt $p$ from the RealToxicityPrompts dataset $\mathcal{D}$, and $\sigma(\cdot)$ is the `toxicity' classifier likelihood.

The BCF treats the toxicity classifier as a binary classifier with threshold $0.5$, and reports the fraction of $k$ completions that score higher than the threshold -- a measure of the frequency of completions that feature `toxicity'.
%; $\text{BCF} \triangleq \mathbb{E}_{p \in \mathcal{D}} \left[ \frac{1}{k} \sum_{i=1}^k \mathbb{1} \left( \sigma(c_i(p)) > 0.5 \right) \right]$.
% \begin{align}
%     \text{BCF} &\triangleq \mathbb{E}_{p \in \mathcal{D}} \left[
%         \frac{1}{k}
%         \sum_{i=1}^k
%         \mathbb{1} \left( \sigma(c_i(p)) > 0.5 \right)
%     \right].
% \end{align}
We report both EML and BCF as percentages $\in [0, 100]$, where 0 is the best performance.

\subsection{Farrell Lexicon}

We also evaluate models using the Farrell lexicon -- a recent misogyny-specific benchmark developed by subject-matter experts \cite{FarrellExploring2019,FarrellUse2020}.
This open-source framework was designed for studying misogynistic discourse in online communities such as Reddit, and includes nine individual lexicons for characterising nine different types of misogyny (Physical Violence, Sexual Violence, Hostility, Patriarchy, Stoicism, Racism, Homophobia, Belittling, and Flipped Narrative).

To evaluate completions against this benchmark, we perform whole-word matching against lexicon terms by splitting completions into words at punctuation and whitespace characters, and considering one or more matches as evidence for that type of misogyny (\ie{} the lexicon for each type of misogyny acts as a binary classifier).\footnotemark{}
For each completion, we then take a boolean `OR' over the nine individual misogyny categories to compute a result for `Any' type of misogyny.
We also report the BCF for the Farrell lexicon as a percentage $\in [0, 100]$, where 0 is the best performance.

\footnotetext{%
    N.b. Sub-word matching is inappropriate for this benchmark due to the presence of very short words in some lexicons (\eg{} `re') which trigger many false positives.
}

\section{Comparison of the Perspective API and Detoxify toxicity classifiers}
\label{app:detoxify-perspective-comparison}

We were unable to secure a quota increase for the Jigsaw Perspective API toxicity classifier -- meaning our full suite of evaluations would have taken months using the free rate-limited quota of 1 query per second.
Instead of sub-sampling our evaluation datasets, we opted to substitute with the open-source Detoxify classifier \cite{HanuDetoxify2020}.

We use the `multilingual' version of this model, which is an \verb|xlm-roberta-base| transformer fine-tuned on the data provided by Jigsaw for the \textit{Multilingual Toxic Comment Classification} challenge (a combination of Wikipedia comments and and CivilComments \cite{DucheneBenchmark2023} dataset).
It scores very competitively on the Kaggle leaderbord for this task ($92.11\%$, best is $95.36\%$), and we manually verify agreement with the Perspective scores by computing Pearson's correlation coefficient over the prompts and continuations in the  from the RealToxicityPrompts dataset.

\begin{figure*}[t]
    \centering
    \begin{subfigure}[b]{0.33\textwidth}
        \centering
        \includegraphics[width=\textwidth]{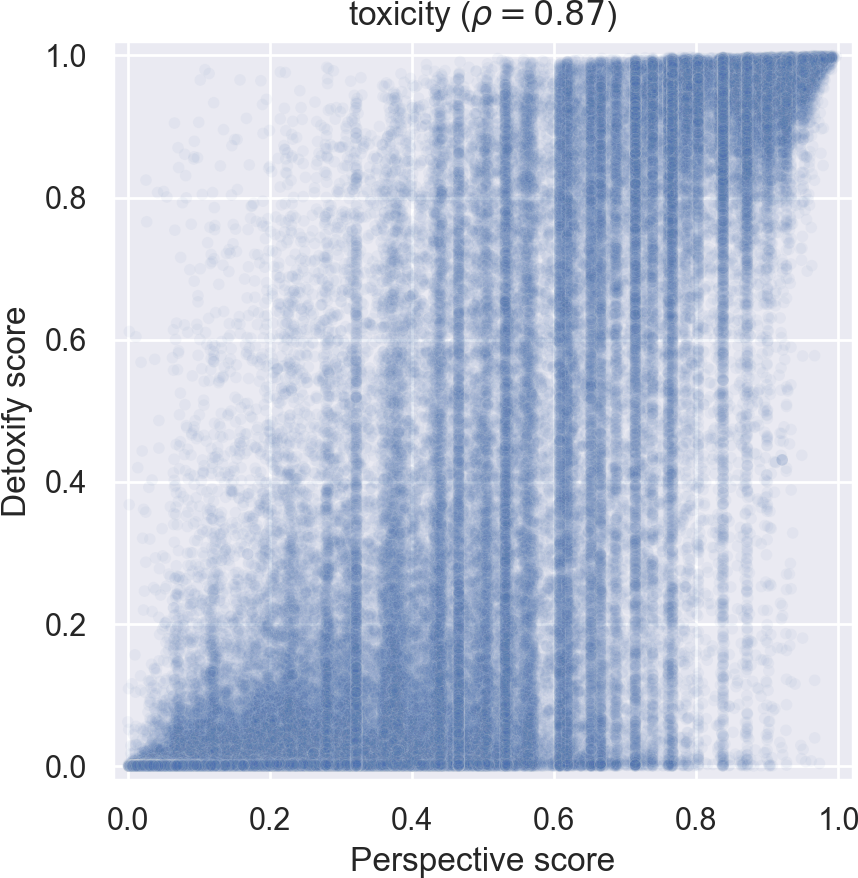}
        %\caption{Foo}
        %\label{fig:foo}
    \end{subfigure}%
    \hfill
    \begin{subfigure}[b]{0.33\textwidth}
        \centering
        \includegraphics[width=\textwidth]{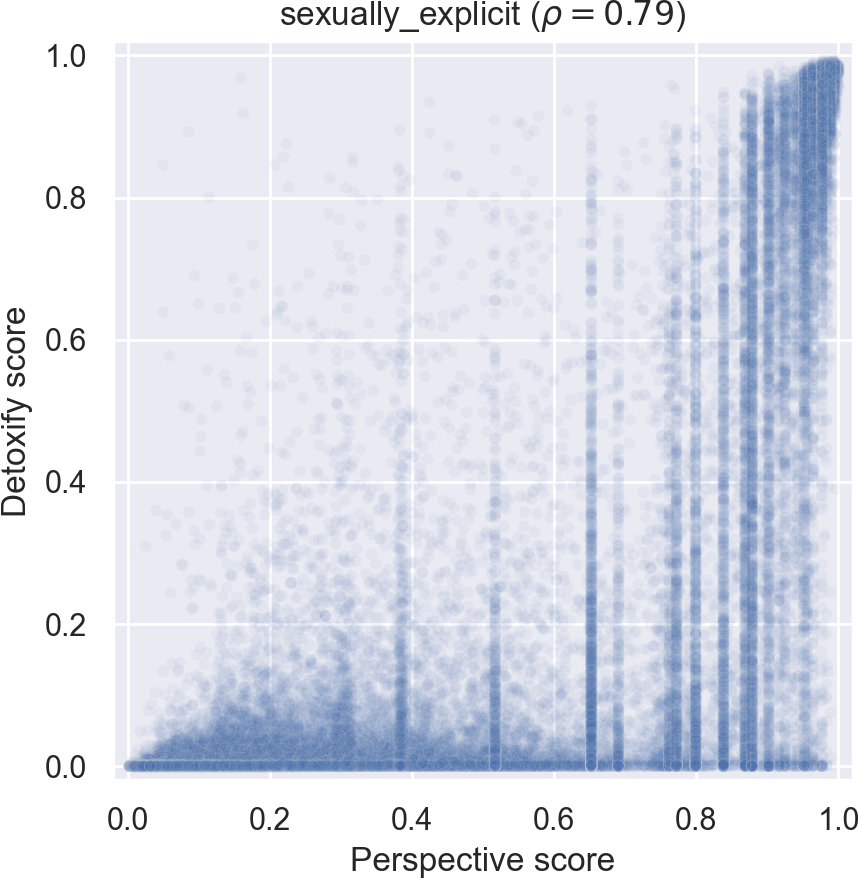}
        %\caption{Foo2}
        %\label{fig:foo2}
    \end{subfigure}%
    \hfill
    \begin{subfigure}[b]{0.33\textwidth}
        \centering
        \includegraphics[width=\textwidth]{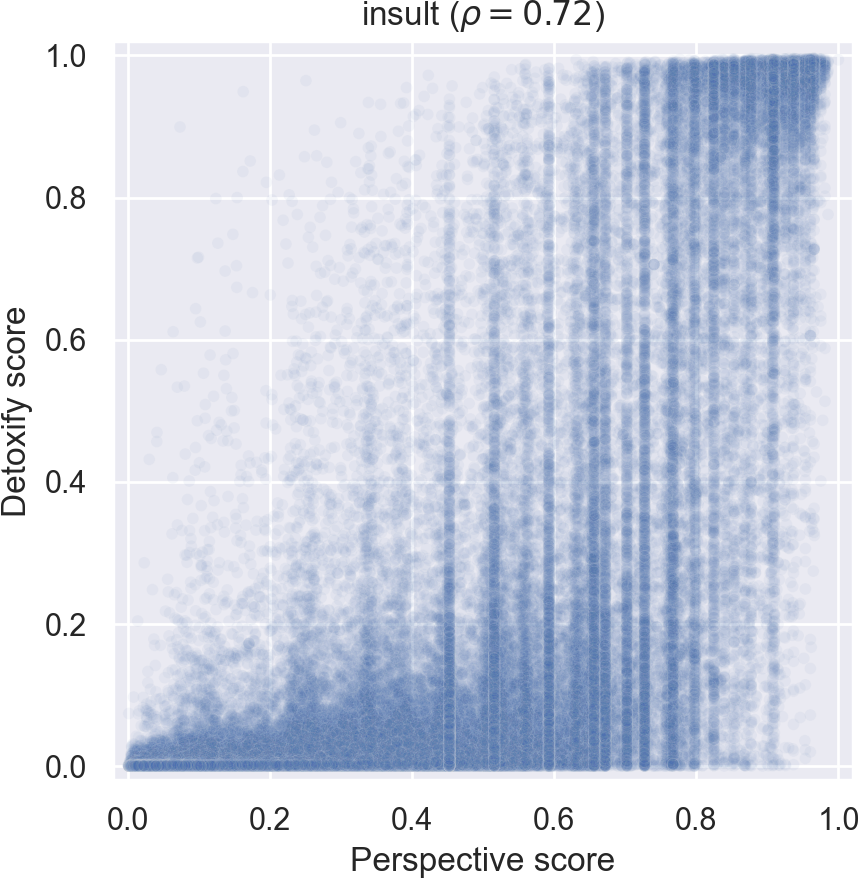}
        %\caption{Foo3}
        %\label{fig:foo3}
    \end{subfigure}%
    \vspace{10pt}
    \begin{subfigure}[b]{0.33\textwidth}
        \centering
        \includegraphics[width=\textwidth]{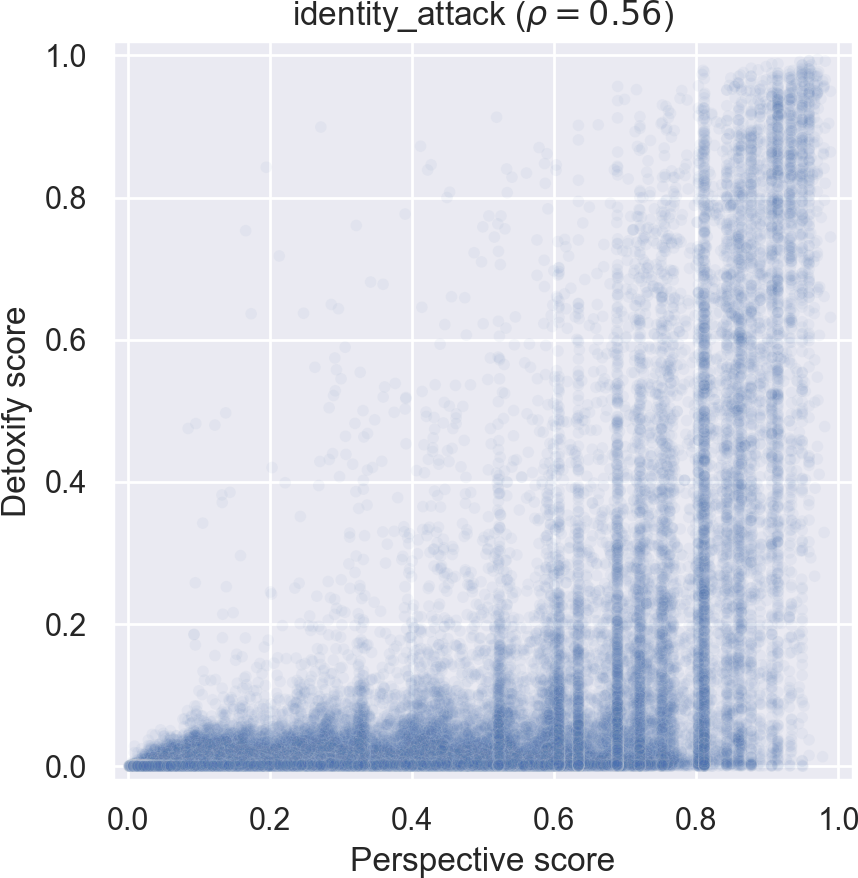}
        %\caption{Foo}
        %\label{fig:foo}
    \end{subfigure}%
    \hfill
    \begin{subfigure}[b]{0.33\textwidth}
        \centering
        \includegraphics[width=\textwidth]{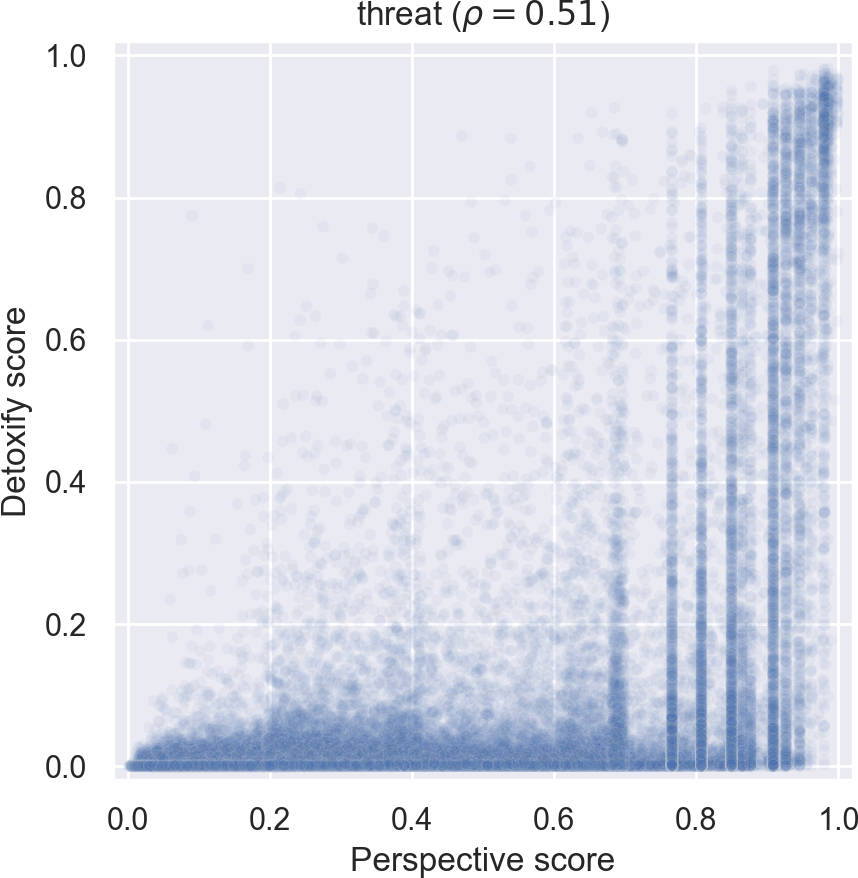}
        %\caption{Foo2}
        %\label{fig:foo2}
    \end{subfigure}%
    \hfill
    \begin{subfigure}[b]{0.33\textwidth}
        \centering
        \includegraphics[width=\textwidth]{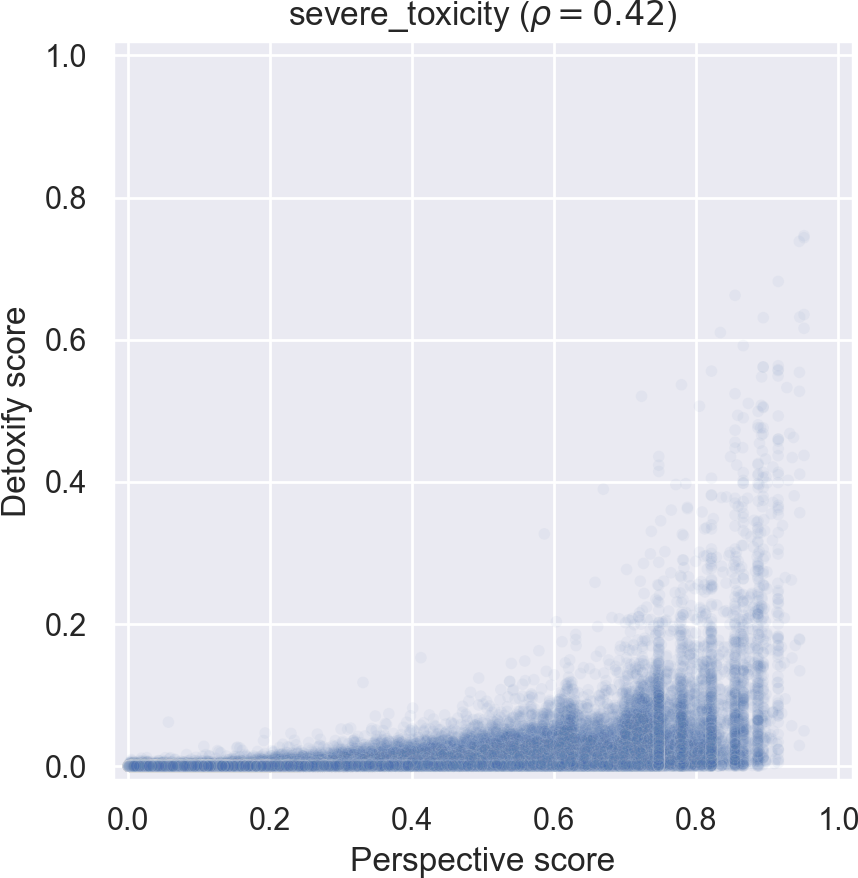}
        %\caption{Foo3}
        %\label{fig:foo3}
    \end{subfigure}%
    \caption{%
        Scatter plot showing Pearson's coefficient of correlation for Perspective API scores and Detoxify-multilingual scores on prompts and completions from the RealToxicityPrompts dataset.
        First row, left to right; `toxicity' ($\rho=0.86$); `sexually\_explicit ($\rho=0.79$); `insult' ($\rho=0.72$).
        Second row, left to right; `identity\_attack' ($\rho=0.56$); `threat' ($\rho=0.51$); `severe\_toxicity' ($\rho=0.42$).
    }
    \label{fig:perspective-detoxify-scatter}
\end{figure*}

\Cref{fig:perspective-detoxify-scatter} shows scatter-plots for each of the classifier models output by both Detoxify and Perspective API on the prompts and completions from the RealToxicityPrompts dataset.
For each plot, we also compute the Pearson correlation coefficient to verify how correlated these classifiers are.
From best to worst, the model correlations are `toxicity' ($\rho=0.86$), `sexually\_explicit ($\rho=0.79$), `insult' ($\rho=0.72$), `identity\_attack' ($\rho=0.56$), `threat' ($\rho=0.51$), and `severe\_toxicity' ($\rho=0.42$).
The Detoxify-multilingual model does not include models equivalent to Perspective API's `flirtation' or `profanity', so we exclude these items.

In general, we observe that Detoxify tends to assign lower scores than the scores recorded in the RealToxicityPropmts from the Perspective API, however all models are positively correlated.

We note also that the scores recorded in the RealToxicityPrompts have been shown in recent scholarship to have shifted to lower values in general when the prompts are re-evaluated with the latest version(s) of the Perspective API \cite{YeNoisyHate2023}, which suggests that the current version of Perspective API may actually have higher correlation with Detoxify-multilingual than what our results suggest.

\clearpage
\section{Additional benchmark results}
\label{app:additional-evaluation-results}

We include results for additional classifier models from both Detoxify-multilingual, and the Farrell lexicon.

\begin{table*}[h]
\caption{Detoxify-multilingual -- the expected maximum likelihood score when doing $k=25$ completions per prompt in RealToxicityPrompts (lower is better). Values are mean and $95\%$ confidence interval over 25 completions over the whole dataset. Best value(s) are \textbf{bold}, worst are \textit{italic}.}
\label{tab:detoxify-prompted-eml}
\resizebox{\textwidth}{!}{%
\begin{tabular}{@{}rccccccc@{}}
\toprule
\bfseries Model & \bfseries toxicity & \bfseries severe\_toxicity & \bfseries obscene & \bfseries identity\_attack & \bfseries insult & \bfseries threat & \bfseries sexual\_explicit \\
\midrule
Pre-trained & 49.35 $\pm$ 0.24 & 1.49 $\pm$ 0.03 & 17.66 $\pm$ 0.20 & 9.69 $\pm$ 0.12 & 25.42 $\pm$ 0.21 & 11.08 $\pm$ 0.13 & 17.30 $\pm$ 0.20 \\
\midrule
Incels & {\cellcolor[HTML]{D9E2F3}} \bfseries 32.05 $\pm$ 0.24 & {\cellcolor[HTML]{D9E2F3}} \bfseries 0.69 $\pm$ 0.02 & {\cellcolor[HTML]{D9E2F3}} \bfseries 13.28 $\pm$ 0.18 & {\cellcolor[HTML]{D9E2F3}} \bfseries 3.80 $\pm$ 0.08 & {\cellcolor[HTML]{D9E2F3}} \bfseries 11.29 $\pm$ 0.14 & {\cellcolor[HTML]{D9E2F3}} \bfseries 2.38 $\pm$ 0.06 & {\cellcolor[HTML]{D9E2F3}} \bfseries 10.12 $\pm$ 0.15 \\
ForeverAlone & {\cellcolor[HTML]{FBE4D5}} 76.71 $\pm$ 0.18 & {\cellcolor[HTML]{FBE4D5}} 2.59 $\pm$ 0.03 & {\cellcolor[HTML]{FBE4D5}} 49.57 $\pm$ 0.25 & 9.71 $\pm$ 0.12 & {\cellcolor[HTML]{FBE4D5}} 40.13 $\pm$ 0.21 & {\cellcolor[HTML]{D9E2F3}} 10.57 $\pm$ 0.12 & {\cellcolor[HTML]{FBE4D5}} 32.52 $\pm$ 0.22 \\
\bottomrule
\end{tabular}%
}
\end{table*}

\begin{table*}[h]
\caption{Detoxify-multilingual -- the binary classification probability when doing $k=25$ completions per prompt in RealToxicityPrompts (lower is better). Values are mean and $95\%$ confidence interval over 25 completions over the whole dataset. Best value(s) are \textbf{bold}, worst are \textit{italic}.}
\label{tab:detoxify-prompted-bcp}
\resizebox{\textwidth}{!}{%
\begin{tabular}{@{}rccccccc@{}}
\toprule
\bfseries Model & \bfseries toxicity & \bfseries severe\_toxicity & \bfseries obscene & \bfseries identity\_attack & \bfseries insult & \bfseries threat & \bfseries sexual\_explicit \\
\midrule
Pre-trained & \bfseries 6.65 $\pm$ 0.08 & 0.01 $\pm$ 0.00 & \bfseries 1.59 $\pm$ 0.04 & 0.48 $\pm$ 0.02 & \bfseries 1.81 $\pm$ 0.03 & 0.60 $\pm$ 0.02 & \bfseries 1.93 $\pm$ 0.05 \\
\midrule
Incels & {\cellcolor[HTML]{FBE4D5}} 12.03 $\pm$ 0.13 & 0.01 $\pm$ 0.00 & {\cellcolor[HTML]{FBE4D5}} 5.13 $\pm$ 0.08 & {\cellcolor[HTML]{FBE4D5}} 0.80 $\pm$ 0.03 & {\cellcolor[HTML]{FBE4D5}} 2.86 $\pm$ 0.06 & {\cellcolor[HTML]{D9E2F3}} 0.48 $\pm$ 0.03 & {\cellcolor[HTML]{FBE4D5}} 3.42 $\pm$ 0.07 \\
ForeverAlone & {\cellcolor[HTML]{FBE4D5}} 9.08 $\pm$ 0.06 & {\cellcolor[HTML]{D9E2F3}} \bfseries 0.00 $\pm$ 0.00 & {\cellcolor[HTML]{FBE4D5}} 3.84 $\pm$ 0.04 & {\cellcolor[HTML]{D9E2F3}} \bfseries 0.33 $\pm$ 0.01 & {\cellcolor[HTML]{FBE4D5}} 2.19 $\pm$ 0.02 & {\cellcolor[HTML]{D9E2F3}} \bfseries 0.38 $\pm$ 0.01 & {\cellcolor[HTML]{FBE4D5}} 2.02 $\pm$ 0.03 \\
\bottomrule
\end{tabular}%
}
\end{table*}

\begin{table*}[h]
\caption{Farrell lexicon -- the percentage frequency of misogynistic generation when doing $k=25$ completions per prompt in RealToxicityPrompts (lower is better). Values are mean and $95\%$ confidence interval over the whole dataset. Best value(s) are \textbf{bold}, worst are \textit{italic}.}
\label{tab:farrell-prompted}
\resizebox{\textwidth}{!}{%
\begin{tabular}{@{}rcccccccccc@{}}
\toprule
\bfseries Model & \bfseries Any & \bfseries Belittling & \bfseries Flipping N. & \bfseries Homophobia & \bfseries Hostility & \bfseries Patriarchy & \bfseries Physical V. & \bfseries Racism & \bfseries Sexual V. & \bfseries Stoicism \\
\midrule
Pre-trained & 5.72 $\pm$ 0.05 & \bfseries 0.52 $\pm$ 0.01 & \bfseries 0.03 $\pm$ 0.00 & 0.04 $\pm$ 0.00 & \bfseries 1.14 $\pm$ 0.02 & \bfseries 0.02 $\pm$ 0.00 & 3.01 $\pm$ 0.04 & 0.77 $\pm$ 0.02 & 0.25 $\pm$ 0.01 & \bfseries 0.10 $\pm$ 0.01 \\
\midrule
Incels & {\cellcolor[HTML]{FBE4D5}} 9.40 $\pm$ 0.11 & {\cellcolor[HTML]{FBE4D5}} 1.57 $\pm$ 0.05 & {\cellcolor[HTML]{FBE4D5}} 0.55 $\pm$ 0.03 & {\cellcolor[HTML]{FBE4D5}} 0.08 $\pm$ 0.01 & {\cellcolor[HTML]{FBE4D5}} 2.08 $\pm$ 0.05 & \bfseries 0.02 $\pm$ 0.00 & {\cellcolor[HTML]{D9E2F3}} 2.57 $\pm$ 0.06 & {\cellcolor[HTML]{FBE4D5}} 1.19 $\pm$ 0.04 & {\cellcolor[HTML]{D9E2F3}} 0.16 $\pm$ 0.01 & {\cellcolor[HTML]{FBE4D5}} 1.67 $\pm$ 0.05 \\
ForeverAlone & {\cellcolor[HTML]{D9E2F3}} \bfseries 5.50 $\pm$ 0.04 & {\cellcolor[HTML]{FBE4D5}} 0.77 $\pm$ 0.01 & {\cellcolor[HTML]{FBE4D5}} 0.13 $\pm$ 0.00 & {\cellcolor[HTML]{D9E2F3}} \bfseries 0.02 $\pm$ 0.00 & {\cellcolor[HTML]{FBE4D5}} 1.51 $\pm$ 0.02 & {\cellcolor[HTML]{FBE4D5}} 0.03 $\pm$ 0.00 & {\cellcolor[HTML]{D9E2F3}} \bfseries 2.47 $\pm$ 0.03 & {\cellcolor[HTML]{D9E2F3}} \bfseries 0.49 $\pm$ 0.02 & {\cellcolor[HTML]{D9E2F3}} \bfseries 0.04 $\pm$ 0.00 & {\cellcolor[HTML]{FBE4D5}} 0.16 $\pm$ 0.01 \\
\bottomrule
\end{tabular}%
}
\end{table*}

% \clearpage
% \section{Example model generations}
% \label{app:example-model-generations}

% Below, we include the 15 prompt-completion pairs from each model that escalate from a non-misogynistic prompt to a misogynistic completion with the highest frequency.
% For each completion, we include the corresponding prompt from the RealToxicictyPrompts dataset, and highlight the completion in {\color{red} \textbf{red boldface font}}.

% \vspace{20pt}

% {%
%     \color{red}%
%     \textbf{%
%         \danger%
%         Content warning: we include the prompts and completions without editing or censoring.
%         Reader discretion is advised.
%         \danger%
%     }%
% }

% \clearpage

% \hl{TODO} generate misogyny escalation appendix for inclusion.

% \input{appendices/app-escelations}

\end{document}